\documentclass[pdflatex,sn-mathphys-num]{sn-jnl}

\usepackage{multirow}%
\usepackage{amsmath,amssymb,amsfonts}%
\usepackage{amsthm}%
\usepackage{mathrsfs}%
\usepackage[title]{appendix}%
\usepackage{xcolor}%
\usepackage{textcomp}%
\usepackage{manyfoot}%
\usepackage{booktabs}%
\usepackage{algorithm}%
\usepackage{algorithmicx}%
\usepackage{algpseudocode}%
\usepackage{listings}%
\usepackage{float}
\usepackage{graphicx}%

\theoremstyle{thmstyleone}%
\theoremstyle{thmstyletwo}%

\theoremstyle{thmstylethree}%

\begin{document}

\title[PUMA Challenge 2024]{Cracking the PUMA Challenge in 24 Hours with CellViT++ and nnU-Net}

\author[1]{\fnm{Negar} \sur{Shahamiri}}\email{Negar.Shahamiri@uk-essen.de}
\author[1,2]{\fnm{Moritz} \sur{Rempe}}\email{Moritz.Rempe@uk-essen.de}
\author[1,2]{\fnm{Lukas} \sur{Heine}}\email{Lukas.Heine@uk-essen.de}
\author[1,2,3,4]{\fnm{Jens} \sur{Kleesiek}}\email{Jens.Kleesiek@uk-essen.de}
\author[1,2]{\fnm{Fabian} \sur{Hörst}}\email{Fabian.Hoerst@uk-essen.de}

\affil[1]{\orgdiv{Institute for Artificial Intelligence in Medicine (IKIM)}, \orgname{University Hospital Essen (AöR)}, \orgaddress{\street{Girardetstraße 2}, \city{Essen}, \postcode{45131}, \state{NRW}, \country{Germany}}}

\affil[2]{\orgdiv{Cancer Research Center Cologne Essen (CCCE), West German Cancer Center Essen}, \orgname{University Hospital Essen (AöR)}, \orgaddress{\street{Hufelandstraße 55}, \city{Essen}, \postcode{45147}, \state{NRW}, \country{Germany}}}

\affil[3]{\orgdiv{German Cancer Consortium (DKTK, Partner site Essen)}, \orgaddress{\city{Heidelberg}, \postcode{69120}, \state{Baden-Württemberg}, \country{Germany}}}

\affil[4]{\orgdiv{Department of Physics}, \orgname{TU Dortmund University}, \orgaddress{\street{Otto-Hahn-Straße 4}, \city{Dortmund}, \postcode{44227}, \state{NRW}, \country{Germany}}}


\abstract{Automatic tissue segmentation and nuclei detection is an important task in pathology, aiding in biomarker extraction and discovery. The panoptic segmentation of nuclei and tissue in advanced melanoma (PUMA) challenge aims to improve tissue segmentation and nuclei detection in melanoma histopathology. Unlike many challenge submissions focusing on extensive model tuning, our approach emphasizes delivering a deployable solution within a 24-hour development timeframe, using out-of-the-box frameworks. The pipeline combines two models, namely $\text{CellViT}^{{\scriptscriptstyle ++}}$ for nuclei detection and nnU-Net for tissue segmentation.  Our results demonstrate a significant improvement in tissue segmentation, achieving a Dice score of 0.750, surpassing the baseline score of 0.629. For nuclei detection, we obtained results comparable to the baseline in both challenge tracks. The code is publicly available at \url{https://github.com/TIO-IKIM/PUMA}. 
}


\maketitle

\section{Introduction}\label{sec1}
Melanoma is a highly aggressive form of skin cancer that is treatable in its early stages. Once melanoma reaches an advanced stage, it requires immune checkpoint inhibition therapy, with responses observed only in 50\% of the patients. This underscores the importance of prognostic biomarkers, such as tumor-infiltrating lymphocytes (TILs). Unfortunately, manual assessment of TILs is time-consuming and costly. To advance research, the PUMA Challenge was initiated, with the goal of segmenting connected tissue regions and isolated nuclei in hematoxylin- and eosin-stained tissue sections~\citep{schuiveling2024novel}.\\
We propose a deep learning-based pipeline consisting of two main pathways: (1) panoptic nuclei segmentation using $\text{CellViT}^{{\scriptscriptstyle ++}}$~\citep{horst2025cellvit++} and (2) tissue segmentation using nnU-Net~\citep{Isensee2020}. 
The focus of our approach differs from many challenge submissions: Instead of trying to win the challenge using ``tweaks'' such as extensive model testing and task-specific model architectures, model ensembling, and metric optimization (based on the test dataset distribution), our goal was to provide a reasonable good solution limited resource setting by using out-of-the-box frameworks. In particular, we set ourselves the goal to provide a solution within 24 hours of development time. During this timeframe, we prepared the datasets for $\text{CellViT}^{{\scriptscriptstyle ++}}$ and nnU-Net, installed the frameworks and environments, defined and implemented the training strategy, and prepared the Docker container template, including the inference scripts. In fact, we were able to train the final $\text{CellViT}^{{\scriptscriptstyle ++}}$ model within the 24-hour window. However, due to the time-consuming nature of training nnU-Net, we extended beyond this time frame to complete the tissue segmentation training. Our contributions to this challenge are as follows: 
\begin{enumerate}
    \item \textbf{Resource constraints}: Our solution is notable for achieving performance within a resource-constrained environment utilizing out-of-the-box frameworks.
    \item \textbf{Performance}: We achieved results comparable to the challenge baseline for nuclei detection using the $\text{CellViT}^{{\scriptscriptstyle ++}}$ approach. By pre-training nnU-Net on additional data, we outperform the tissue segmentation challenge baseline solution by a considerable margin.
\end{enumerate}

\section{Methods}\label{sec2}

\begin{figure}
    \centering
    \includegraphics[width=1\linewidth]{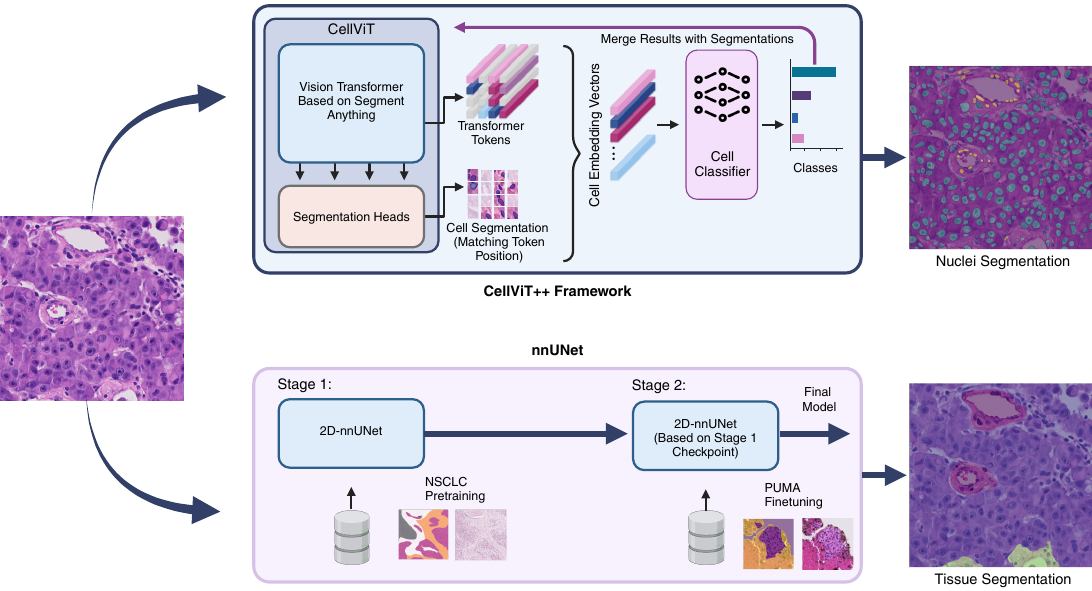}
    \caption{Pipeline overview consisting of  $\text{CellViT}^{{\scriptscriptstyle ++}}$ for nuclei detection and nnU-Net with 2-stage training for tissue segmentation. Created in BioRender~\citep{biorender_figure}.}
    \label{fig:overview}
\end{figure}

\subsection{PUMA Dataset and Evaluation}
The PUMA dataset includes 155 primary and 155 metastatic melanoma regions of interest ($1024\times1024\,\text{px}$) with annotations for both nuclei and tissue~\citep{schuiveling2024novel}. Within the public training dataset, there are 206 images (50/50 split), along with a preliminary test set of ten images. The concealed final test set comprises 94 samples. The competition offers two separate tracks, distinguished by their classification strategies for nuclei. Track 1 categorizes into three types: tumor cells, TILs, and other cells. In contrast, Track 2 expands the classification to ten categories (Tab.~\ref{tab3}). Both tracks employ the same classes for tissue segmentation: tumor, stroma, necrosis, epidermis, and blood vessels. The assessment uses the F1 score for nuclei detection and the Dice score for tissue segmentation. For reference, the challenge baseline includes HoVer-Next for nuclei and nnU-Net for tissue.

\subsection{Pipeline}
Figure~\ref{fig:overview} shows an overview of our pipeline. The $\text{CellViT}^{{\scriptscriptstyle ++}}$ model is used without any modifications compared to the original publication~\citep{horst2025cellvit++}. In this setup, the segmentation backbone (CellViT~\citep{cellvit}), is not trained. Instead, we only trained the lightweight cell classifiers for the nuclei taxonomies corresponding to track 1 and track 2, respectively. We used nnU-Net with the same setup as the challenge baseline for tissue segmentation. However, we pre-trained the network using a publicly available histopathological dataset. The pre-training dataset, published by~\citet{Kludt2024}, consists of 2000 samples ($512 \times 512\,\text{px}$) from patients with non-small cell lung cancer (NSCLC), which contain 12 annotated tissue classes. Both frameworks, $\text{CellViT}^{{\scriptscriptstyle ++}}$ and nnU-Net, are available as out-of-the-box solutions, requiring minimal setup effort and only basic deep learning experience for implementation. We utilized the frameworks without significant modifications and did not adopt the models.

\subsection{Training Setup}
We trained a separate classifier for each track for the $\text{CellViT}^{{\scriptscriptstyle ++}}$ model. We fine-tuned the classifiers and performed a hyperparameter sweep, each with 100 experiments, to determine the optimal settings. The training dataset was divided into 154 training, 26 validation, and 26 test images, with early stopping applied on the validation set. The final model for submission was selected based on test results from the internal test set. Each sweep took approximately eight hours of computation time. 

For tissue segmentation, we pre-trained the nnU-Net network on the NSCLC dataset using one fold (1600 training, 400 validation images), selecting the best checkpoint based on validation performance. As the NSCLC samples are only partially annotated, we implemented a custom trainer with masked cross-entropy loss to ignore empty areas. We then fine-tuned the pre-trained model on the PUMA dataset using the challenge setup, with a batch size 12, and selected the best checkpoint for submission. We used a single fold rather than nnU-Net’s default model ensemble. All training configurations are available in the GitHub repository.

\section{Results}\label{sec3}
The results on the preliminary test set for our method and the challenge baseline are presented in Tab.~\ref{tab1} and~\ref{tab2}. While our approach achieves performance comparable to the baseline for nuclei detection across both tracks (Tab.~\ref{tab3}), the pre-training strategy for nnU-Net significantly surpasses the PUMA-only baseline. Notably, the class-wise comparison in Tab.~\ref{tab2} demonstrates that pre-training enables accurate segmentation of necrotic regions, a tissue class the baseline model failed entirely.

\begin{table}[t]
\centering
\begin{tabular}{lcccccccccccc}
    \toprule
    \textbf{Method}      & & & \multicolumn{2}{c}{\textbf{Track 1}}   & & & \multicolumn{2}{c}{\textbf{Track 2}} \\ \midrule
                & & & Tissue (Dice) $\uparrow$   & Nuclei (F1) $\uparrow$ & & & Tissue (Dice) $\uparrow$ & Nuclei (F1) $\uparrow$ \\ \midrule
    Baseline    & & & 0.629           & 0.638       & & & 0.629         & 0.227       \\
    Ours        & & & 0.750           & 0.611       & & & 0.750         & 0.226       \\ \bottomrule
\end{tabular}
\caption{Comparison of tissue segmentation and nuclei detection}
\label{tab1}
\vspace{-3mm}
\end{table}

\begin{table}[t]
    \centering
    \begin{tabular}{lcccccc}
        \toprule
        \textbf{Method} & \textbf{Tumor} & \textbf{Stroma} & \textbf{Necrosis} & \textbf{Epidermis} & \textbf{Blood vessel} & \textbf{Average}\\
        \midrule
        Baseline & 0.934 & 0.849 & 0.000 & 0.853 & 0.508 & 0.629 \\
        Ours  & 0.913 & 0.826 & 0.845 & 0.792 & 0.372 & 0.750 \\
        \bottomrule
    \end{tabular}
    \caption{Detailed comparison of tissue segmentation performance using Dice-score}
    \label{tab2}
    \vspace{-3mm}
\end{table}
\vspace{-5mm}

\section{Discussion and Conclusion}\label{sec4}
In this work, we presented a deep learning-based pipeline for the PUMA Challenge, focusing on achieving competitive performance under time constraints using out-of-the-box frameworks. Our approach leveraged $\text{CellViT}^{{\scriptscriptstyle ++}}$ for nuclei detection and nnU-Net for tissue segmentation, with the only notable modification being the pre-training of nnU-Net on an external histopathological dataset. Overall, our work highlights the potential of leveraging pre-trained models and out-of-the-box frameworks to achieve robust segmentation performance within limited development time and resources. We believe that this approach can facilitate the deployment of automated histopathological analysis pipelines in resource-constrained environments, advancing melanoma research and clinical practice.

\backmatter

\bmhead{Acknowledgments}
This work received funding from \lq KITE' (Plattform für KI-Translation Essen) from the REACT-EU initiative (\url{https://kite.ikim.nrw/}, EFRE-0801977) and the Cancer Research Center Cologne Essen (CCCE). 
Moritz Rempe and Jens Kleesiek received funding from the Bruno \& Helene Jöster Foundation. The authors acknowledge that this manuscript was edited with the assistance of LLMs. The authors declare no competing interests. The technical implementation of this work was a collaborative effort by all authors.

\bibliography{sn-bibliography}

\appendix

\setcounter{table}{0}
\renewcommand{\thetable}{A.\arabic{table}}

\section*{Appendix}

\begin{table}[!h]
\setlength{\belowcaptionskip}{-3mm}
\centering
\footnotesize
\begin{tabular}{llcccc}
\toprule
                    &                   & \multicolumn{2}{c}{\textbf{Track 1}} & \multicolumn{2}{c}{\textbf{Track 2}} \\ \toprule
\textbf{Supercategory} & \textbf{Subcategory} & \textbf{Baseline}   & \textbf{Ours}  & \textbf{Baseline}   & \textbf{Ours}  \\ \midrule
Tumor               & Tumor             & 0.734               & 0.725          & 0.720               & 0.709          \\ \midrule
Lymphocytes         & Lymphocytes       & 0.753               & 0.744          & 0.692               & 0.703          \\
                    & Plasma            &                     &                & 0.073               & 0.053          \\ \midrule
Other               & Histiocytes       & 0.424               & 0.364          & 0.192               & 0.202          \\
\textbf{}           & Neutrophils       &                     &                & 0.000               & 0.000          \\
\textbf{}           & Stromal           &                     &                & 0.224               & 0.241          \\
\textbf{}           & Epithelium        &                     &                & 0.025               & 0.080          \\
\textbf{}           & Endothelium       &                     &                & 0.084               & 0.018          \\
                    & Apoptotic         &                     &                & 0.100               & 0.047          \\ \midrule
Average             & \textbf{}         & 0.637               & 0.611          & 0.227               & 0.226          \\ \bottomrule
\end{tabular}%
\caption{Detailed comparison of nuclei detection performance using F1-score}
\label{tab3}
\end{table}

\end{document}